\documentclass[nohyperref]{article}

\usepackage{microtype}
\usepackage{graphicx}
\usepackage{subfigure}
\usepackage{booktabs} 

\usepackage{hyperref}

\usepackage{import}
\usepackage{enumitem}
\usepackage{mathtools}
\usepackage{bbm}
\usepackage{lipsum}
\usepackage{xparse}
\usepackage{tabularx}
\usepackage{multirow}
\usepackage[frozencache,cachedir=.]{minted}
\usepackage{xcolor}
\usepackage{listings}
\usepackage{adjustbox}

\newcommand{\prs}[1]{\left( #1 \right)}

\newcommand{\set}[1]{\{{#1}\} }

\NewDocumentCommand{\LeftComment}{s m}{%
  \Statex \IfBooleanF{#1}{\hspace*{\ALG@thistlm}}\(\triangleright\) #2}

\definecolor{codegreen}{rgb}{0,0.6,0}
\definecolor{codegray}{rgb}{0.5,0.5,0.5}
\definecolor{codepurple}{rgb}{0.58,0,0.82}
\definecolor{backcolour}{rgb}{0.95,0.95,0.92}

\lstdefinestyle{mystyle}{
    backgroundcolor=\color{backcolour},   
    commentstyle=\color{codegreen},
    keywordstyle=\color{magenta},
    numberstyle=\tiny\color{codegray},
    stringstyle=\color{codepurple},
    basicstyle=\ttfamily\footnotesize,
    breakatwhitespace=false,         
    breaklines=true,                 
    captionpos=b,                    
    keepspaces=true,                 
    numbers=left,                    
    numbersep=5pt,                  
    showspaces=false,
    morekeywords={self},
    showstringspaces=false,
    showtabs=false,                  
    tabsize=2
}

\usepackage[accepted]{icml2022}

\usepackage{amsmath}
\usepackage{amssymb}
\usepackage{mathtools}
\usepackage{amsthm}

\usepackage[capitalize,noabbrev]{cleveref}

\theoremstyle{plain}

\theoremstyle{definition}

\theoremstyle{remark}

\usepackage[disable,textsize=tiny]{todonotes}

\icmltitlerunning{Pure Noise to the Rescue of Insufficient Data}

\begin{document}

\twocolumn[
\icmltitle{Pure Noise to the Rescue of Insufficient Data:\\
Improving Imbalanced Classification by Training on Random Noise Images}



\icmlsetsymbol{equal}{*}

\begin{icmlauthorlist}
\icmlauthor{Shiran Zada}{yyy}
\icmlauthor{Itay Benou}{yyy}
\icmlauthor{Michal Irani}{yyy}
\end{icmlauthorlist}

\icmlaffiliation{yyy}{Department of Computer Science and Applied Mathematics, Weizmann Institute of Science, Rehovot, Israel}

\icmlcorrespondingauthor{Shiran Zada}{shiran.elyahuzada@weizmann.ac.il}
\icmlcorrespondingauthor{Itay Benou}{itay.benou@weizmann.ac.il}
\icmlcorrespondingauthor{Michal Irani}{michal.irani@weizmann.ac.il}

\icmlkeywords{Machine Learning, ICML}

\vskip 0.3in
]



\printAffiliationsAndNotice{}  

\begin{abstract}

    Despite remarkable progress on visual recognition tasks, deep neural-nets still struggle to generalize well when training data is scarce or highly imbalanced, rendering them \mbox{extremely} vulnerable to real-world examples. In this paper, we present a surprisingly simple yet highly effective method to mitigate this limitation: \textbf{using pure noise images as additional training data}. Unlike the common use of additive noise or adversarial noise for data augmentation, we propose an entirely different perspective by directly training on pure random noise images.
    We present a new Distribution-Aware Routing Batch Normalization layer (DAR-BN), which enables training on pure noise images in addition to natural images within the same network. 
    This encourages generalization and suppresses overfitting. 
    Our proposed method significantly improves imbalanced classification performance, 
 obtaining state-of-the-art results on a large variety of long-tailed image classification datasets (CIFAR-10-LT, CIFAR-100-LT, ImageNet-LT, Places-LT, and CelebA-5). 
    Furthermore, our method is extremely simple and easy to use as {a general new augmentation tool} (on top of existing augmentations), and can be incorporated in any training scheme. It does not require any specialized data generation or training procedures, thus keeping training fast and efficient.
    \vspace*{-0.3cm}

\end{abstract}

\section{Introduction}
\label{sec:intro}
Large-scale annotated datasets play a vital role in the success of deep neural networks for visual recognition tasks. While popular benchmark datasets are usually well-balanced (e.g., CIFAR~\cite{krizhevsky2009learning}, Places~\cite{zhou2017places}, ImageNet~\cite{deng2009imagenet}), data in the real world often follows a \emph{long-tail distribution}. Namely, most of the data belongs to several \emph{majority classes}, while the rest is spread across a large number of \emph{minority classes}~\cite{buda2018systematic, reed2001pareto, liu2019large}. 
Training on such imbalanced datasets results in models that are biased towards majority classes, demonstrating poor generalization on minority classes. There are two common approaches to compensate for class imbalance during training: (i)~\textit{re-weighting} the loss term so that prediction errors on minority samples are given higher penalties~\cite{huang2016learning, cui2019class, hong2021disentangling, ren2020balanced}, and (ii)~\textit{resampling} the dataset to re-balance the class distribution during training~\cite{chawla2002smote, kim2020m2m, mullick2019generative}.
This can be done by under-sampling majority classes~\cite{drummond2003c4}, or by over-sampling of minority classes~\cite{shen2016relay, haixiang2017learning, kang2019decoupling}.

However, \textit{re-weighting} methods typically suffer from overfitting the minority classes~\cite{kim2020m2m}. \textit{Resampling} techniques also suffer from well-known limitations: under-sampling majority classes may impair classification accuracy due to loss of information, while over-sampling leads to overfitting on minority classes~\cite{buda2018systematic}. Several methods have been proposed to alleviate these limitations, including augmentation-based methods~\cite{chu2020feature, mullick2019generative, liu2020deep}, learning from majority classes~\cite{kim2020m2m} and evolutionary under-sampling~\cite{galar2013eusboost}.

The data scarcity in minority classes thus  poses a very challenging problem~\cite{kim2020m2m}. This is especially true in highly imbalanced datasets, where minority classes contain very few samples (e.g., 5 images per class, vs. thousands of images in majority classes). In such cases, overfitting is almost inevitable, even with extensive data augmentation, since the ability to produce a significant variety of new observations from just a few  samples is extremely limited. 

\begin{figure*}[t]
\vspace*{-0.2cm}
    \centering
    \begin{minipage}[c]{0.68\linewidth}
    \includegraphics[page=1,width=\textwidth]{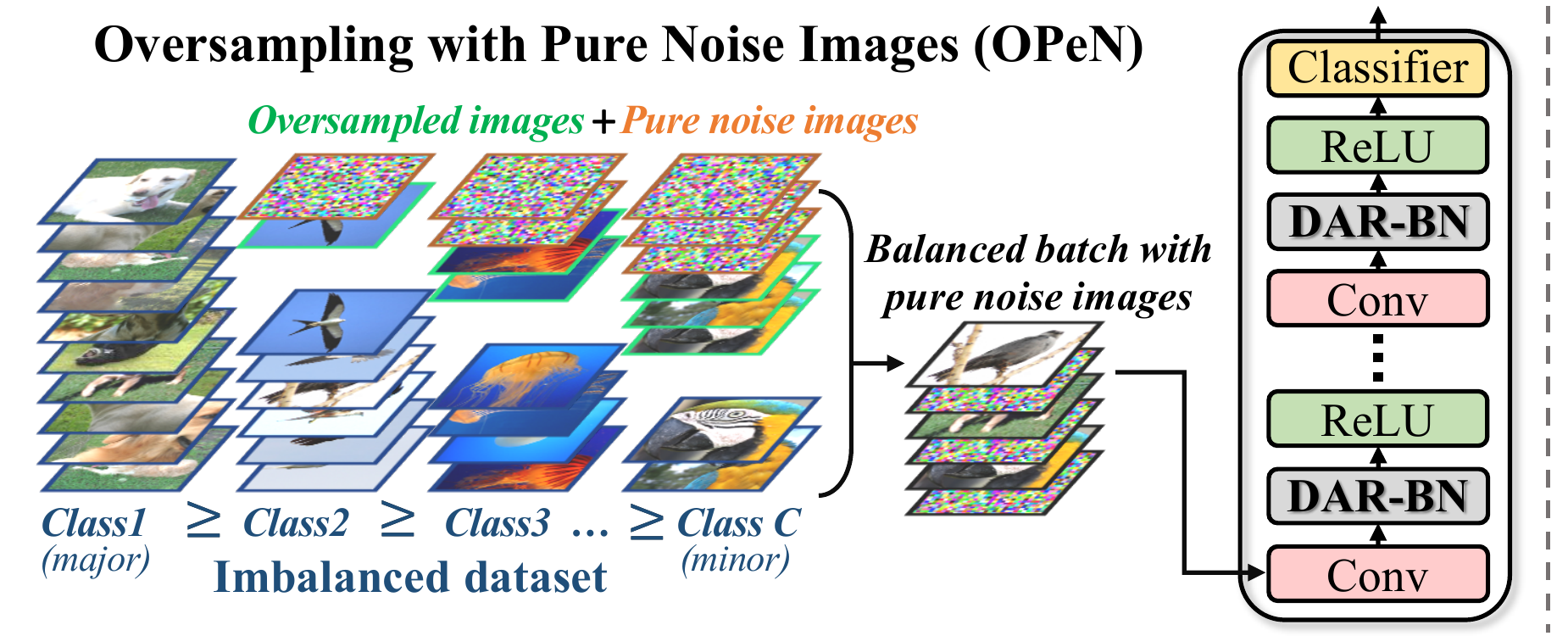}
    \end{minipage}
     \hfill
    \begin{minipage}[c]{0.30\linewidth}
    \includegraphics[page=1,width=\textwidth]{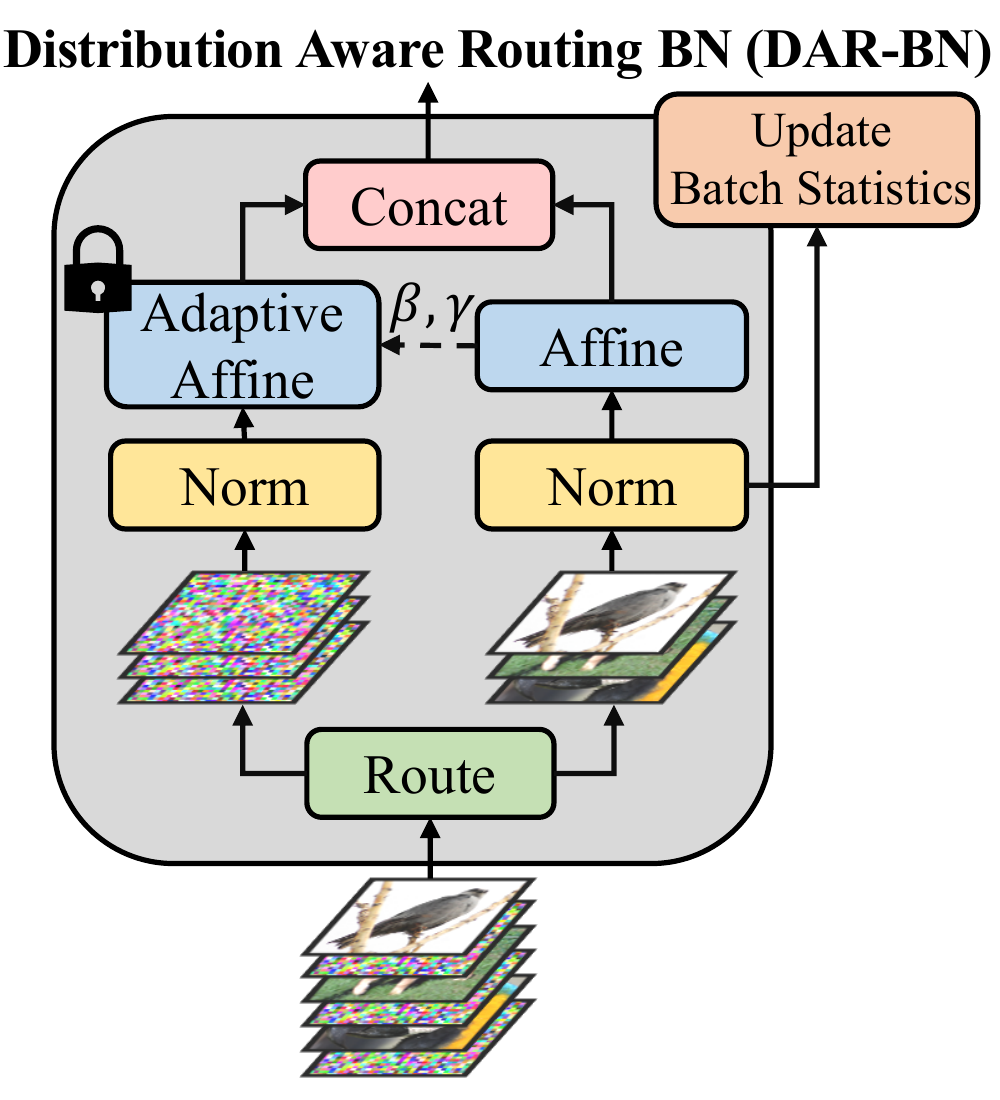}
    \end{minipage}
    \vspace*{-0.3cm}
    {\caption{\textbf{Method overview}. \  {\it \textbf{(Left)} 
    OPeN re-balances an imbalanced dataset with pure-noise images, in addition to oversampled natural images. In OPeN, we replace the standard Batch Normalization layer with DAR-BN. \textbf{(Right)}  ``Distribution Aware Routing BN'' (DAR-BN) handles the distribution gap between  natural images and pure-noise images, by normalizing them separately. The affine parameters learned on the natural input only, are used to correctly scale and shift the noise input.}}    \label{fig:methodOverview}}
    \vspace*{-0.5cm}

\end{figure*}

\begin{figure}
  \begin{minipage}[c]{0.47\columnwidth}
    \includegraphics[width=4cm]{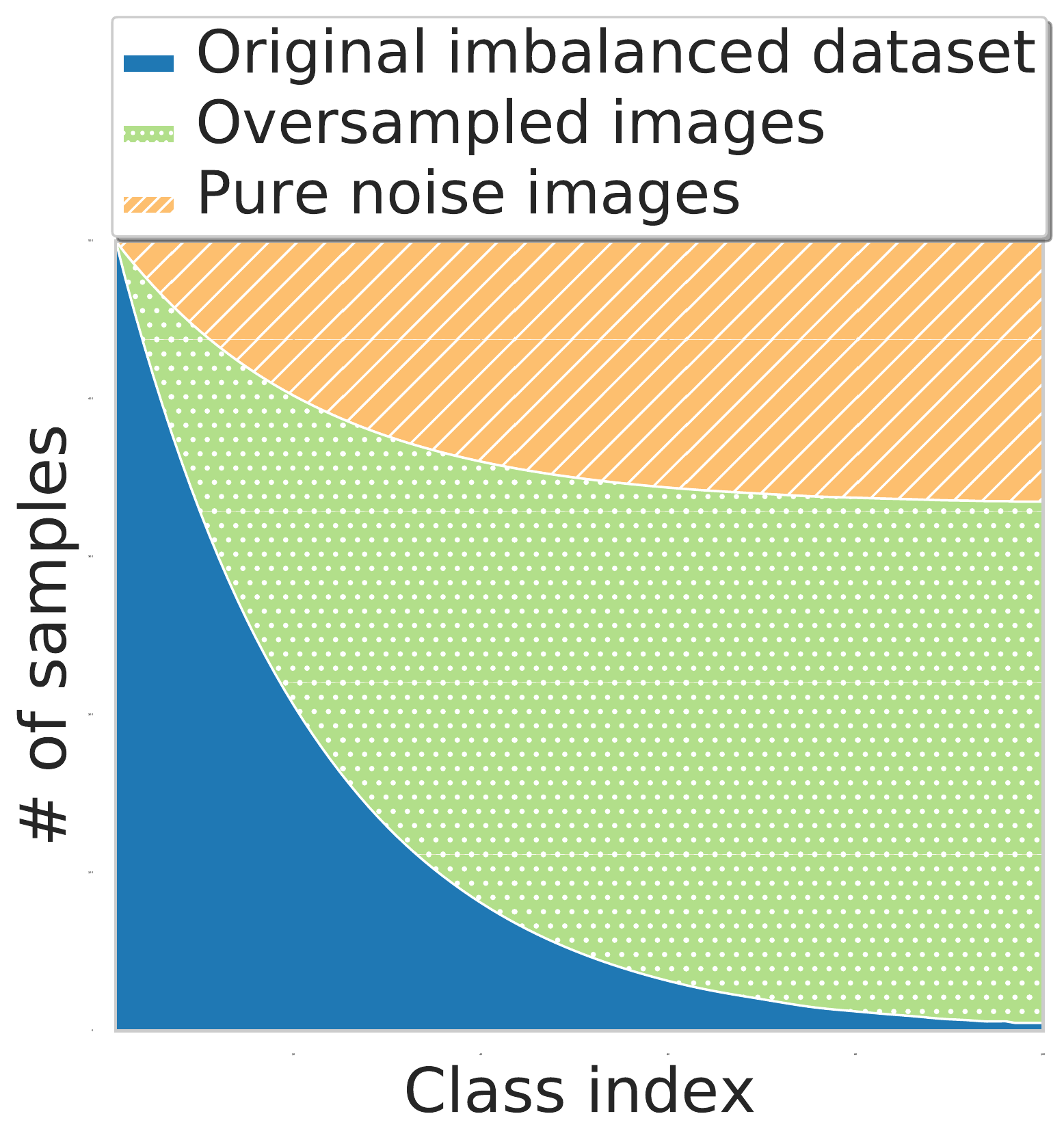}
  \end{minipage}\hfill
  \begin{minipage}[c]{0.47\columnwidth} 
    \vspace*{-0.6cm}
    \caption{\textbf{Re-balanced dataset with OPeN}. {\it To balance the dataset, the original images (blue) are oversampled (green) together with additional pure random noise images (orange). The amount of pure noise added to each class is inversely proportional to its size. 
    }
\label{fig:rebalanced_dataset}
    } 
  \end{minipage}
  \vspace*{-0.8cm}
\end{figure}

In this work, we directly address this problem by taking a new perspective on data re-balancing for imbalanced classification. Unlike traditional resampling approaches, we do not restrict ourselves to training strictly on existing images and their augmentations, thus bypassing this limitation. Specifically, we propose \emph{\textbf{generating pure random noise images and using them as additional training data}} (especially for minority classes). We show that training on pure noise images can suppress overfitting and encourage generalization, leading to state-of-the-art results on commonly used imbalanced classification benchmarks (\cref{sec:experiments}). We further provide an intuitive explanation as to why this counter-intuitive approach works in practice (\cref{sec:intuition}).
To facilitate learning on pure noise images, which are out-of-distribution of natural images, we present a new batch normalization layer called \textit{Distribution-Aware Routing Batch Normalization} (DAR-BN). Unlike standard Batch-Normalization (BN)~\cite{ioffe2015batch} that assumes that all inputs are drawn from the same distribution, DAR-BN is specifically designed to mitigate the distribution shift between two different input domains (namely, natural images and pure noise images).

We note that many previous works have used noise as a
form of data augmentation to improve the accuracy and robustness of deep learning models~\cite{koziarski2017image, lopes2019improving}. These methods,
however, show limited improvement when the training data
is scarce~\cite{koziarski2017image}. 
This is due to the fact that applying small doses of additive/multiplicative noise to existing images produces samples in close vicinity to the original ones, thus limiting the data variability. Adding large amounts of noise, on the other hand, degrades models' performance due to the large distribution shift from natural images (\cref{appendix:additive_noise}).
In contrast, in our method -- \textbf{OPeN} (\textbf{O}versampling with \textbf{P}ur\textbf{e} \textbf{N}oise Images), 
the model is explicitly trained on pure
noise images that are far-off the natural images manifold, while explicitly handling the distribution shift, thus promoting generalization.

\noindent \underline{Our contributions are therefore several fold:}
\vspace{-0.1cm}
\begin{itemize}[topsep=0pt,itemsep=-1ex,partopsep=1ex,parsep=1ex,leftmargin=*]
\item \emph{State-of-the-art results} on multiple imbalanced classification benchmarks: CIFAR10-LT and CIFAR100-LT~\cite{cao2019learning}, ImageNet-LT~\cite{liu2019large}, Places-LT~\cite{liu2019large}, CelebA-5~\cite{kim2020m2m}.
\item To our best knowledge, we are the first to successfully use pure-noise images for training deep image recognition models. We provide extensive empirical evidence to its \mbox{improved generalization capabilities (and intuition why).}
\item We introduce a new \emph{distribution-aware batch normalization layer} (DAR-BN), that can bridge the distribution-gap between different input domains of neural networks. While in this work we used DAR-BN to bridge the gap between real and pure-noise images, it can be applied as a general BN layer for handling any pair of different input domains.
\item Our method is extremely simple to use as \emph{a general new augmentation tool} (on top of existing augmentations), and can be incorporated in any training scheme. It does not require any specialized data generation or training procedures, thus training is fast and efficient. 
Our code  is available at \hyperlink{https://github.com/shiranzada/pure-noise}{https://github.com/shiranzada/pure-noise}.
\end{itemize}

\vspace{-0.15cm}
\section{Related Work}
\vspace{-0.05cm}
\noindent \textbf{\underline{Imbalanced Classification:}} \\
\noindent\textit{Data resampling:} Most data-based approaches for imbalanced classification aim to re-balance the dataset such that minority and majority classes are equally represented during training. This can be achieved by either over-sampling minority classes~\cite{chawla2002smote, wang2014hybrid, shen2016relay} or under-sampling majority classes~\cite{drummond2003c4,liu2008exploratory, galar2013eusboost}. More recent works address class re-balancing using GANs~\cite{mullick2019generative, hao2020annealing} and semi-supervised learning~\cite{wei2021crest}. An oversampling framework related to our work is \textit{M2m}~\cite{kim2020m2m}, in which majority samples are “transferred” to minor classes using adversarial noise perturbations. Our method also belongs to the data re-balancing category, however, it does so by adding pure random noise images as additional training samples rather than using additive noise augmentations. We also note that our method does not require any optimized data creation procedure or using an auxiliary classifier, which allows a simple and efficient training process ($\times 10$ faster than~\cite{kim2020m2m}, with better results).
Our OPeN framework belongs to the category of data resampling approaches, which are therefore most relevant to our work. 

\noindent \textit{Loss re-weighting}: aims to compensate for data imbalance by adjusting the loss function, e.g., by assigning minority samples with higher loss weights than majority samples~\cite{li2021autobalance, buda2018systematic,cui2019class,ren2018learning,park2021influence}. Recently,  \textit{BALMS}~\cite{ren2020balanced} and \textit{LADE}~\cite{hong2021disentangling} both suggested calibrating the predicted logits according to a prior distribution, by adjusting the softmax function or by adding a regularization term to the loss, respectively. 

\noindent \textit{Margin loss}: using a loss function that pushes the decision boundary further away from minority classes samples~\cite{zhang2017range,dong2018imbalanced}. For example, ~\cite{cao2019learning} presented the Label Distribution Aware Margin (\textit{LDAM}) loss, which is combined with deferred re-weighting (DRW) training schedule for improved results. 

\noindent \textit{Decoupled training}: A recent line of work showing that separating the feature representation learning from the final classification task can be beneficial for imbalanced classification~\cite{kang2019decoupling,zhou2020bbn,wang2021contrastive}. E.g., the recently proposed \textit{MiSLAS} method~\cite{zhong2021improving} suggested using a shifted batch normalization layer between the two stages of the decoupling framework, in addition to calibrating the final model predictions using a label-aware smoothing scheme.

\noindent \textbf{\underline{Noise-Based Augmentation:}} \\
Augmenting training data with additive or multiplicative noise has long been in use for training visual recognition models~\cite{holmstrom1992using, bengio2011deep, ding2016convolutional}. The main motivation behind such augmentation techniques is improving the model robustness to noisy inputs and 
preventing its fixation
on specific input features by randomly “occluding” parts of them~\cite{lopes2019improving}. While 
demonstrating some success in reducing overfit~\cite{zur2009noise}, these methods usually provide limited improvement to deep models as they tend to overfit to the specific type of noise used during training~\cite{yin2015noisy}.

Another group of methods that uses additive noise are adversarial training techniques, which aim to “fool” a deep model by perturbing images with small, optimized noise~\cite{goodfellow2014explaining, kurakin2016adversarial}. Particularly relevant is \textit{M2m}~\cite{kim2020m2m}, which  suggests using adversarial noise to "transfer" images from major classes to minor classes in an imbalanced classification setting. Similarly, \textit{AdvProp}~\cite{xie2020adversarial} suggests utilizing adversarial examples for improving accuracy and robustness in a general (balanced) classification setting. 
They try to bridge the distribution gap between two types of inputs (real and adversarial images), for which they use an auxiliary batch normalization layer. 
In our work, however, the training data is enriched using pure noise images rather than adversarial examples. Additionally, \textit{AdvProp} learns two completely separate sets of batch-norm parameters while in our proposed \textit{DAR-BN} the affine parameters are learned only based on real images, and then applied to both data sources.  

\noindent \textbf{\underline{Normalization Layers:}} \\
Since the introduction of batch-normalization~\cite{ioffe2015batch}, various extensions have been proposed to further improve normalization within deep networks, including layer-norm~\cite{ba2016layer}, instance-norm~\cite{ulyanov2016instance} and group-norm~\cite{wu2018group}. In common to all these layers is that they normalize activation maps based on a single set of statistics (i.e., mean and variance) for the entire training set. While this may work well when all data samples are from the same underlying distribution, it is sub-optimal when the data is multi-modal or originates from several different domains~\cite{xie2020adversarial}. Several recent works which relate to ours have addressed this issue: adaptive instance-normalization~\cite{huang2017arbitrary} was introduced for style transfer by adjusting the statistics of content and style inputs. 
Similarly,~\cite{li2018adaptive, xie2020adversarial} propose mitigating the domain shift by keeping separate sets of normalization terms for different domains. In our proposed \textit{DAR-BN} layer we also normalize real and noise inputs using their internal statistics. However, unlike~\cite{li2018adaptive,xie2020adversarial},
DAR-BN uses the affine parameters learned by the \textit{natural activation maps} in order to scale and shift the \textit{noise activation maps}. 
Finally, since at test time inputs are sampled only from the natural images domain, DAR-BN updates the batch statistics only using activation maps of natural images.
We found those difference to be critical to the results.

\vspace{-0.1cm}
\section{Imbalanced Classification using OPeN}
\vspace{-0.05cm}

Figures~\ref{fig:methodOverview} and~\ref{fig:rebalanced_dataset} provide a schematic overview of our approach for imbalanced  classification, which is detailed next.
$\mathcal{D} \coloneqq \bigcup\limits_{i\in [C]}\set{(x_j, c_i)}_{j=1}^{n_i}$ is a long-tailed imbalanced dataset containing $C$ classes $\set{1,2\ldots,C}$, where each class $c_i$ consist of $n_i$ training samples. For simplicity we assume $n_1\ge n_2\ge\cdots \ge n_C$ and $n_1\gg n_C$. In some cases, the ratio between the largest and the smallest class is a factor of 1000. While the training-set  $\mathcal{D}$ is \textit{class-imbalanced}, the test-set is \textit{class-balanced}, and therefore classification of minority classes (with only few samples) is of equal importance to that of majority classes. To compensate for the lack of training data in minority classes, we adopt an oversampling approach that levels the number of samples in each class. However, in contrast to common oversampling techniques~\cite{chawla2002smote, kim2020m2m}, whose training images are solely based on original ones (i.e., their duplications and augmentations), we propose to use also \emph{\textbf{pure random Gaussian noise images as additional training samples}}. As shown in Fig.~\ref{fig:rebalanced_dataset}, for each class $c_i$, we balance the data by adding $n_i^*$$=$$n_{max}$$-$$n_i$ new training images (where $n_{max}$$=$$n_1$ is the largest class), out of which $\delta$$\cdot$$n_i^*$ are pure noise images, and $(1$$-$$\delta)$$\cdot$$n_i^*$ are real (oversampled) images. 
During training, we feed the network with mixed batches containing both natural images (with augmentations) and pure noise images. In Sec.~\ref{sec:OPeN} we further elaborate on this process.
Since pure-noise images are out-of-distribution of natural images, we normalize them separately using a new distribution-aware normalization layer. This is explained in \cref{sec:DAR-BN}. Finally,  Sec.~\ref{sec:intuition}  provides intuition why this improves accuracy of imbalanced classification, especially on minority classes.

\subsection{Oversampling with Pure Noise Images (OPeN)}\label{sec:OPeN}

We define the \emph{representation-ratio} of each class $c_i$ in $\mathcal{D}$ as $\rho_i \coloneqq n_i/n_{max}$.
By definition, minority classes have a smaller representation ratio than majority classes. 
Since standard oversampling results in overfitting of minority classes~\cite{buda2018systematic}, we replace part of the oversampled images with pure random noise images, with the following probability:
\vspace*{-0.2cm}
\begin{equation}\label{eq:probability_to_replace_with_noise}
	\mathbb{P}(\text{replace $x$ with $x_{noise}$} | c_i) = (1-\rho_i)\cdot\delta
\end{equation}
where $c_i$ is the associated class label of image $x$, $\rho_i$ is the representation ratio of class $c_i$, and $x_{noise}$ is randomly sampled from a normal distribution using the mean and variance of all images in the dataset. $\delta\in [0,1]$ is a hyper-parameter defining the ratio between pure noise images and natural images. 
Each class in the dataset has a different number of samples, hence is prone to overfitting to a different extent. 
\cref{eq:probability_to_replace_with_noise} adjusts the number of noise images added per class accordingly. 
Lower $\rho_i$ results in a higher probability to replace a sample from class $c_i$ with a pure random noise image, and vice versa for larger $\rho_i$.

The pure random noise images are generated as follows.
Let $\mathcal{X}$ be the set of training images in the dataset $\mathcal{D}$. We first compute the mean and standard deviation for each color channel $l\in \set{1,2,3}$:
\vspace*{-0.1cm}
\begin{equation}
	\mu_{\mathcal{D},l} = \mathbb{E}(\mathcal{X}[l,\;:\;])\;\;,\;\;
	\sigma_{\mathcal{D},l} = \sqrt{Var(\mathcal{X}[l,\;:\;])}
\end{equation}
Noise images are then sampled 
from he following normal distribution and clipped to the feasible domain [0,1]:
\vspace*{-0.2cm}
\begin{equation}\label{eq:x_noise}
	\hat{x}_{noise} \sim \mathcal{N}\prs{\mu_\mathcal{D}, \sigma_\mathcal{D}}
\end{equation}
\vspace*{-0.4cm}
\begin{equation}\label{eq:x_noise_clamped}
	x_{noise} = \min\prs{\max\prs{\hat{x}_{noise}, 0},1}
\end{equation}
At every epoch, we randomly sample new noise images, as this helps the network to avoid overfitting to specific noise images. 
A pseudo-code for OPeN is shown in \cref{alg:noise_oversample}. Full PyTorch code is provided in~\cref{sec:code}.
\vspace*{-0.2cm}
\begin{algorithm}
	\caption{\mbox{ \hspace{-0.2em}Oversampling with Pure Random Noise \hspace{-0.1em}{\small (OPeN)}}}
	\begin{algorithmic}
		\STATE \textbf{Input}: (i) Imbalanced dataset: $\mathcal{D} = \hspace{-0.5em}\bigcup\limits_{i\in [C]}\set{(x_j, c_i)}_{j=1}^{n_i}$\\ (ii) noise ratio: $\delta\in [0,1]$; (iii) dataset statistics:$\mu_\mathcal{D}$, $\sigma_\mathcal{D}$
		\\\hrulefill
		\STATE Initialize $\rho = \set{\rho_1,\ldots,\rho_C}\coloneqq \set{n_i/n_{max}}_{i=1}^C$
		\STATE $L_D \leftarrow$ Balanced loader for $\mathcal{D}$ using oversampling
		\STATE $B=\set{(x_j,y_j)}_{j=1}^{batch\_size}\leftarrow$ Sample a batch from $L_D$
		\FORALL {$(x_j,y_j) \in B$}
		\STATE $\triangleright$ Compute probability $\xi$ of replacing $x_j$ with noise
		\STATE $\xi\leftarrow (1-\rho[y_j])\cdot\delta$
		\STATE $r\sim Bernoulli(\xi)$
		\IF{$r == 1$}
		\STATE $x_{noise} \sim \mathcal{N}\prs{\mu_\mathcal{D}, \sigma_\mathcal{D}}$
		\STATE $x_j \leftarrow \min\prs{\max\prs{x_{noise}, 0},1}$
		\ENDIF
		\ENDFOR
		\STATE \textbf{return} $B$
	\end{algorithmic}
	\label{alg:noise_oversample}
\end{algorithm}

\vspace*{-0.5cm}
\subsection{\mbox{\bf {\normalsize Distribution-Aware Routing Batch Norm (DAR-BN)}}}\label{sec:dar_bn}
\label{sec:DAR-BN}
Standard Batch Normalization~\cite{ioffe2015batch} is designed to handle the change in the distribution of inputs to layers in deep neural networks, also known as internal covariate shift. However, it assumes that all input samples are taken from the same or similar distributions. Therefore, when inputs originate from several different distributions, BN fails to properly normalize them~\cite{he2019data, xie2020adversarial, xie2019intriguing}. In our framework, OPeN uses pure random noise as additional training examples, which are clearly out of the distribution of natural images. As a result, the layer's input consists of activation maps obtained both from natural images (where the similar distribution assumption holds) as well as pure noise images (whose distribution is very different from that of the natural images in the train or test datasets). We experimentally observe that using noise images with the standard BN layer, leads to a significant degradation in classification results (see \cref{sec:ablation_study_observations}), even below the baseline of not using noise at all. 
This may further suggest why this simple idea (of adding pure noise images as additional training examples) has not been previously proposed as a general tool to improve generalization of deep neural networks. 
To handle the significant distribution gap between random noise images and natural images, we introduce a new normalization layer called ``DAR-BN": \mbox{\textit{Distribution-Aware Routing Batch Normalization}}.

We start by revisiting the standard Batch Normalization (BN)~\cite{ioffe2015batch}, and then explain how we extend it to our proposed DAR-BN. Let $X\in \mathbb{R}^{N\times d\times H\times W}$ denote an input to the normalization layer, where $N$ is the batch size, $H$$\times$$W$ is the spatial dimension size, and $d$ is the number of channels. The BN layer acts on each channel independently by first normalizing the input across the spatial and the batch dimensions, then applying an affine layer with trainable parameters. Formally, for each channel $j\in [d]$:
\vspace*{-0.2cm}
\begin{equation}\label{eq:BN-normalize-train}
	\hat{x}_j = \frac{x_j-\mathbb{E}(x_j)}{\sqrt{Var(x_j)}}
\end{equation}
\vspace*{-0.2cm}
\begin{equation}\label{eq:BN-scale-shift-train}
	{out}_j = \gamma_j \cdot\hat{x}_j -\beta_j
\end{equation}
where  $x_j \in \{x_1,\ldots ,x_d\} \subseteq \mathbb{R}^{N\times H\times W}$ is an input channel (i.e., $x_j=X[:,\;j,\;:,\;:]$) and $\beta_j, \gamma_j$ are trainable parameters per channel. At inference time, the input is normalized using the running mean and running variance that were computed during training using an exponential moving average (EMA) of the batch statistics.
\vspace*{-0.3cm}
\begin{equation}\label{eq:BN-train-statistics-mu}
	\Bar{\mu} = \eta \cdot \mathbb{E}(X)   + (1-\eta) \cdot \Bar{\mu}
\vspace*{-0.4cm}
\end{equation}
\begin{equation}\label{eq:BN-train-statistics-var}
	\Bar{v} = \eta \cdot Var(X)  + (1-\eta) \cdot\Bar{v}
\end{equation}
where $\eta$ is the momentum parameter. Then, at test time the data is normalized by the running mean and variance, i.e., we replace \cref{eq:BN-normalize-train} with:
\vspace*{-0.2cm}
\begin{equation}
	\hspace*{1cm} \hat{x}_j = \frac{x_j-\Bar{\mu}_j}{\sqrt{\Bar{v}_j}}
\end{equation}
To handle the significant distribution shift between random noise images and natural images, we propose DAR-BN, an extension to the standard BN layer. To this goal, DAR-BN normalizes the noise activation maps and the natural activation maps separately. Specifically, assume $X = X_{nat}\cup X_{noise}$ where $X_{nat}, X_{noise}$ are activation maps of natural images and pure noise images in the batch, respectively. DAR-BN replaces \cref{eq:BN-normalize-train} with:
\begin{equation}
	\hat{x}_{nat, j} = \frac{x_{nat, j}-\mathbb{E}(x_{nat, j})}{\sqrt{Var(x_{nat, j})}}
\end{equation}
\vspace{-0.1cm}
\begin{equation}
\hat{x}_{noise,j} = \frac{x_{noise,j}-\mathbb{E}(x_{noise,j})}{\sqrt{Var(x_{noise,j})}}
\end{equation}
Then, motivated by AdaBN~\cite{li2016revisiting} (which is designed to handle the covariate shift for domain adaption/transfer learning), DAR-BN uses the affine parameters learned by the \textit{natural activation maps} in order to scale and shift the \textit{noise activation maps}. Specifically, DAR-BN replaces \cref{eq:BN-scale-shift-train} with:
\vspace*{-0.2cm}
\begin{equation}
	{out}_{nat, j} = \gamma_{\textbf{nat}, j} \cdot\hat{x}_{nat, j} -\beta_{\textbf{nat}, j}
\end{equation}
\vspace*{-0.3cm}
\begin{equation}\label{eq:DAR-BN-scale-shift-train}
	{out}_{noise, j} = \gamma_{\textbf{nat}, j} \cdot\hat{x}_{noise, j} -\beta_{\textbf{nat}, j}
\end{equation}
\Cref{eq:DAR-BN-scale-shift-train} is applied when the parameters $\beta_{nat}, \gamma_{nat}$ remain fixed, such that no update is applied to these parameters in the back-propagation step due to the operation in \cref{eq:DAR-BN-scale-shift-train}.
Finally, since at test time inputs are sampled only from the natural images domain, DAR-BN updates the batch statistics only using activation maps of natural images. Accordingly, equations \cref{eq:BN-train-statistics-mu,eq:BN-train-statistics-var} are replaced with:
\vspace*{-0.2cm}
\begin{equation}
	\Bar{\mu} = \eta \cdot \mathbb{E}(X_{nat})   + (1-\eta) \cdot \Bar{\mu}
\end{equation}
\vspace*{-0.3cm}
\begin{equation}
	\Bar{v} = \eta \cdot Var(X_{nat})  + (1-\eta) \cdot\Bar{v}
\end{equation}
Pseudo-code of DAR-BN is found in \cref{appendix:darbn} (\cref{alg:darbn}) and full PyTorch code  in~\cref{sec:code}.
The distribution shift between pure-noise \& real  activation maps, and its implications on BN, is illustrated in \cref{fig:distribution_shift_ilustraion} (\cref{appendix:darbn}).

\vspace{-0.3em}
\subsection{Underlying Intuition}\label{sec:intuition}
\vspace{-0.3em}
Training directly on random noise images may seem counter-intuitive. However, we claim it provides a unique regularization effect that can significantly improve generalization of minority classes.

Consider the average batch during training in the imbalanced classification setting described above. Each class is represented in the batch according to its relative size in the training set, i.e., its representation ratio, $\rho_i$. When backpropagating, the total gradient can be decomposed into the sum of $C$ individual components, one per class. When applying conventional oversampling (using duplications and augmentations of original images), gradient components of minority classes will increase in magnitude, since they are now over-represented. However, their direction will remain relatively unchanged since oversampled images are usually similar to original ones, thus limiting generalization for these classes. This relates to another well-known drawback of such oversampling methods which tend to perform poorly when the number of samples in minority classes is very small, since the ability to synthesize new and varied samples for those classes is extremely limited~\cite{kim2020m2m}.

\begin{table*}[t]
\vspace{-0.1cm}
\centering
\begin{minipage}[c]{0.6\linewidth}
\vspace*{-0.15cm}
\resizebox{\linewidth}{!}{
\begin{tabular}{l|ccccc}
\toprule
\multirow{2}{*}{Methods} &
  \multicolumn{2}{c}{CIFAR-10-LT} &
  \multicolumn{2}{c}{CIFAR-100-LT} &
  CelebA-5 \\
 &
  {\footnotesize IR=100} &
  {\footnotesize IR=50} &
  {\footnotesize IR=100} &
  {\footnotesize IR=50} &
  {\footnotesize IR=10.7} \\ \midrule
ERM &
  {$79.6${\scriptsize $\pm$0.2}} &
  $84.9${\scriptsize $\pm$0.4} &
  $47.0${\scriptsize $\pm$0.5} &
  $52.4${\scriptsize $\pm$0.4} &
  $78.6$ {\scriptsize $\pm$0.1} \\
Oversampling &
  {$75.1${\scriptsize $\pm$0.4}} &
  $82.2${\scriptsize $\pm$0.4} &
  $42.5${\scriptsize $\pm$0.3} &
  $48.0${\scriptsize $\pm$0.2} &
  $76.4$ {\scriptsize $\pm$0.2} \\
LADM-DRW${}^{\S}$ &
  {$80.5${\scriptsize $\pm$0.6}} &
  $85.3${\scriptsize $\pm$0.2} &
  $46.8${\scriptsize $\pm$0.2} &
  $52.6${\scriptsize $\pm$0.2} &
  $78.5${\scriptsize $\pm$0.5} \\
M2m${}^{\S}$ &
  {$81.3${\scriptsize $\pm$0.4}} &
  $85.5${\scriptsize $\pm$0.3} &
  $46.5${\scriptsize $\pm$0.5} &
  $52.9${\scriptsize $\pm$0.2} &
  $76.9${\scriptsize $\pm$0.4} \\
MiSLAS${}{}^\odot$ &
  {\hspace{-1.6em}82.1} &
  \hspace{-1.6em}$85.7$ &
  \hspace{-1.6em}$47.0$ &
  \hspace{-1.6em}$52.3$ &
  - \\
\textbf{OPeN} &
  {\textbf{84.6}{\scriptsize $\pm$\textbf{0.2}}} &
  \textbf{87.9}{\scriptsize $\pm$\textbf{0.2}} &
  \textbf{51.5}{\scriptsize $\pm$\textbf{0.4}} &
  \textbf{56.3}{\scriptsize $\pm$\textbf{0.4}} &
  \textbf{79.7}{\scriptsize $\pm$\textbf{0.2}} \\ \midrule\midrule
ERM + AA &
  {$81.4${\scriptsize $\pm$0.3}} &
  $86.4${\scriptsize $\pm$0.2} &
  $49.9${\scriptsize $\pm$0.4} &
  $55.7${\scriptsize $\pm$0.4} &
  $79.3${\scriptsize $\pm$0.5} \\
BALMS + AA${}^\odot$ &
  {\hspace{-1.6em}84.9} &
  - &
  \hspace{-1.6em}50.8 &
  - &
  - \\
\textbf{OPeN + AA} &
  {\textbf{86.1}{\scriptsize $\pm$\textbf{0.1}}} &
  \textbf{89.2}{\scriptsize $\pm$\textbf{0.2}} &
  \textbf{54.2}{\scriptsize $\pm$\textbf{0.5}} &
  \textbf{59.8}{\scriptsize $\pm$\textbf{0.5}} &
  \textbf{80.9}{\scriptsize $\pm$\textbf{0.4}} \\ \bottomrule
\end{tabular}}
\vspace{-0.28cm}
\caption{\textbf{Comparisons on CIFAR-10-LT, CIFAR-100-LT, CelebA-5.} {\it  Rows with~$\odot$ denote results  as reported in the original papers. Rows with $\S$ denote results reproduced with the same architecture as in our experiments, for fair comparison (new results are higher than reported in the original papers). AA stands for AutoAugment optimized on CIFAR-10. Missing results indicate datasets not evaluated in the cited papers. \textbf{AA is \underline{not} a legal augmentation for CIFAR-10-LT and ImageNet-LT (as it was optimized on their full balanced datasets). Top part of the table is thus \underline{without} AA. However, since BALMS was trained with AA, we add such a comparison in the bottom part of the table.}
}}
\label{tab:results_cifar_celeb}
\end{minipage}
\hfill
\begin{minipage}[c]{0.38\linewidth}
\centering
\resizebox{0.75\linewidth}{!}{
\begin{tabular}{l|c}
\toprule
Methods  & ImageNet-LT \\ \midrule
ERM                       & 51.1                         \\
Oversampling  \hspace{0.8cm}            & \hspace{0.7cm}49.0\hspace{0.7cm}                         \\
BALMS${}^\diamondsuit$              & 52.1                         \\
LADE${}^\diamondsuit$               & 53.0                         \\
MisLAS${}^{\odot}$ & 52.7                         \\
MisLAS ${}^{\S}$ & 53.7                         \\
\textbf{OPeN}      & \textbf{55.1}                \\ 
\bottomrule
\end{tabular}}
\vspace{-0.2cm}
\caption{\textbf{Comparison on ImageNet-LT} {\it using ResNeXt-50. Rows with $\diamondsuit$ denote results reported in~\cite{hong2021disentangling}. 
The row with $\odot$ denotes the result reported in the original paper (with ResNet-50). 
The row with $\S$ denotes the result reproduced by us with ResNeXt-50. 
OPeN outperforms all previous methods, obtaining state-of-the-art results.}}
\label{tab:results_imagenet}
\vspace{0.2cm}
\begin{minipage}[c]{0.55\linewidth}
\resizebox{\linewidth}{!}{
\begin{tabular}{l|c}
\toprule
Methods                                 & Places-LT
 \\\midrule
ERM                           & 29.9                       \\
Oversampling                                                & 38.1                       \\
BALMS${}^{\odot}$ & 38.7                       \\
LADE${}^{\odot}$                    & 38.8                       \\
MiSLAS${}^{\odot}$                     & 40.4                       \\
\textbf{OPeN}                                        & \textbf{40.5}              \\ 
\bottomrule
\end{tabular}
}
\end{minipage}
\begin{minipage}[c]{0.41\linewidth}
\vspace*{-0.3cm}
\caption{\textbf{Comparison on Places-LT}
{\it using ImageNet pretrained ResNet-152. Rows with $\odot$ denote results reported in the original papers.}}
\label{tab:results_places}
\end{minipage}
\end{minipage}
\vspace{-0.5cm}
\end{table*}

In contrast, using the proposed OPeN resampling scheme alleviates both of these problems: (i)~From a training point of view, oversampling with pure noise images also increases the magnitude of minority gradient components, but at the same time adds stochasticity to their direction. \ On CIFAR-100-LT, the mean gradient magnitude and the direction variance using OPeN are $\times$2 and $\times$9 larger (respectively) than using oversampling \emph{without pure-noise images}. This stochasticity has a regularization effect on the training process, whose strength is \textit{inversely proportional} to the class size. This way, overfitting of minority classes can be suppressed, and generalization is encouraged. (ii)~By using random noise images, generation of new training samples is not limited by the variety of existing samples in the data. This way, we bypass the limitation posed by the small number of minority samples, and explicitly teach the network to handle inputs that are significantly out of its training-set distribution. In particular, the network learns to expect much higher variability and uncertainty in the test images of minority classes. Indeed, at test time, as our experiments suggest, this translates into increased generalization performance. We note that many previous works have used noise as a form of data augmentation. These methods, however, show slight improvement when training data is scarce~\cite{koziarski2017image}, mostly since applying small doses of noise produces new images that are in close vicinity to original ones, thus providing limited data variability.

One can also understand how our method mitigates data imbalance from another perspective. Since noise inputs are completely random and are class independent, they in fact carry no information except for the class labels we assign to them. Consequently, a key effect of using noise images is on the prior class probabilities learnt by the network. Since in the proposed re-sampling scheme more noise images are assigned to minority classes, we hypothesize that the network learns to implicitly encode these prior probabilities and correct its predictions accordingly.

\vspace{-0.2cm}
\section{Experiments:  Imbalanced Classification}\label{sec:experiments}
\vspace{-0.1cm}
We evaluate our method on five  benchmark datasets for imbalanced classification: CIFAR-10-LT, CIFAR-100-LT, ImageNet-LT, Places-LT, and CelebA-5. 
We follow the evaluation protocol used in~\cite{liu2019large, kim2020m2m} for imbalanced classification tasks: 
The model is trained on the \textit{class-imbalanced} training set, but then evaluated on a  \textit{balanced class distribution} test set. 
Our results (summarized in \cref{tab:results_cifar_celeb,tab:results_imagenet,tab:results_places}) exhibit state-of-the-art performance on all these datasets. Our implementation details can be found in \cref{sec:Experimental_Setup}.

\vspace{-0.5em}
\subsection{Experimental Setup}
\textbf{Imbalanced (Long-Tail) Datasets.}
The \emph{Imbalance-Ratio} (IR) of a longtail dataset is defined as IR=$n_{max}/n_{min}$, where $n_{max}$ and $n_{min}$ are the number of training images in its largest and smallest class, respectively. 
The five long-tailed datasets we used are described below, and their information is summarized in \cref{tab:datasets_info} in \cref{sec:Experimental_Setup}.
\begin{itemize}[topsep=0pt,itemsep=-1ex,partopsep=1ex,parsep=1ex,leftmargin=*]
	\item \textbf{CIFAR-10-LT} \textbf{\& CIFAR-100-LT}~\cite{cao2019learning}\textbf{.} The full balanced CIFAR-10/100 datasets~\cite{krizhevsky2009learning} consist of 50,000 training images,
	and 10,000 test images (split uniformly into 10/100 classes, respectively). 
	Their long-tailed versions, CIFAR-10/100-LT 
	\cite{cao2019learning}, were constructed by an exponential decay sampling of the number of training images per class, while their corresponding test sets remain unchanged (i.e., uniform class distribution).  
	We evaluate our method on the  challenging dataset settings (IR=50, IR=100). 
	\item \textbf{ImageNet-LT} \textbf{\& Places-LT}~\cite{liu2019large}\textbf{.} Both datasets 
	are long-tail subset of the original large-scale (balanced)  ImageNet~\cite{deng2009imagenet} and Places~\cite{zhou2017places}. The imbalanced datasets were constructed by sampling the original dataset following the Pareto distribution~\cite{reed2001pareto} with $\alpha$$=$$6$ as its power value. The resulting long-tailed datasets have a smallest class of size 5, and a largest  class of size 1280 for ImageNet-LT (IR=256), and 4980 for Places-LT (IR=996).
	\item \textbf{CelebA-5}~\cite{kim2020m2m}\textbf{.} The standard CelebA dataset~\cite{liu2015faceattributes} consists of face images with 40 binary attributes per image. CelebA-5 dataset was proposed in~\cite{mullick2019generative} by selecting samples from non-overlapping attributes of hair color (\textit{blonde, black, bald, brown,  gray}). Naturally, the resulting dataset is imbalanced (with IR=$10.7$), as human hair colors are not uniformly distributed. The images were then resized to $64$$\times$$64$ pixels. \citet{kim2020m2m} constructed a smaller version of the imbalanced dataset by sampling each class with a ratio 1:20, preserving the IR.
\end{itemize}

	\noindent\textbf{Baseline methods.} On top of comparing to recent leading methods~\cite{cao2019learning,hong2021disentangling,ren2020balanced,zhong2021improving,kim2020m2m}, we also compare ourselves under the same training parameters, augmentations and architectures to the following baselines: (i)~Empirical Risk Minimization (ERM): training without any re-balancing scheme); (ii)~Oversampling: re-balancing the dataset by oversampling minority classes with augmentations; (iii)~ERM + AutoAugment (AA) ~\cite{cubuk2019autoaugment}. 

\subsection{Results}\label{subsec:imbalanced_results}
\Cref{tab:results_cifar_celeb,tab:results_imagenet,tab:results_places} show the results for imbalanced image classification. \textbf{ \emph{OPeN obtains SOTA results on all benchmark datasets.}}  For example, on CIFAR-10-LT and CIFAR-100-LT with an imbalance ratio of 100, OPeN outperforms the previous SOTA method, MiSLAS~\cite{zhong2021improving} \textbf{by 4.5\%} \emph{on both datasets}. On ImageNet-LT, OPeN is higher than previous SOTA method, LADE~\cite{hong2021disentangling}, \textbf{by 2.1\%}. 
On Places-LT, OPeN achieves comparable results (0.1\% better) to previous SOTA, MiSLAS~\cite{zhong2021improving}.
Since our method uses a stronger network (WideResNet-28-10), for fair comparison we reproduced the results of~\cite{cao2019learning,kim2020m2m} with the same architecture.
This improved their results compared to the original papers (Table~\ref{tab:results_cifar_celeb}), yet are still inferior to ours.

The above results were obtained \emph{without} using AutoAugment (AA)~\cite{cubuk2019autoaugment}. 
Using AutoAugment for training extremely small/longtailed subsets of CIFAR and ImageNet is unfair~\cite{azuri2021generative}, since AutoAugment was optimized using the \emph{entire  large and balanced} CIFAR-10 and ImageNet datasets. 
However, since BALMS~\cite{ren2020balanced}  report  results only with AutoAugment, we evaluated our method also with that setting. OPeN with AA outperforms BALMS  \textbf{by 1.1\%} and \textbf{3.4\%} on CIFAR-10-LT and CIFAR-100-LT, respectively. 
We further note that OPeN obtains state-of-the-art results even without using AutoAugment. 

\noindent\textbf{Generalization of minority classes.}
Besides improving the \emph{mean} accuracy (reported in ~\cref{tab:results_cifar_celeb,tab:results_imagenet,tab:results_places}), finer exploration reveals that most of this overall improvement stems from a \emph{dramatic improvement in classification accuracy of minority classes}, while preserving the accuracy of majority classes. Specifically, OPeN improves the accuracy of the 20 smallest classes of {CIFAR100-LT} (with IR=100, where minority classes have 5-12 samples) by 13.9\% above baseline ERM training, \emph{\textbf{from mean accuracy of 11.6\% to 25.5\%}}.  OPeN also outperforms the baseline deferred oversampling~\cite{cao2019learning} (without noise images) by 4.3\% on the same subset of minor classes. On CIFAR10-LT, OPeN improves generalization of the two smallest classes by 6.3\% compared to deferred oversampling, and \emph{\textbf{by 15.6\% above ERM training}}. 
These findings provide empirical evidence to our hypothesis that adding pure noise to minority classes (as opposed to only augmenting the existing training images) significantly diminishes the overfitting problem and increases the generalization capabilities. Please see \cref{appendix:Generalization_class_size} for more detailed evaluations.

\begin{figure}[t]
\vspace*{-0.1cm}
	\centering
    \includegraphics[page=1,width=\linewidth]{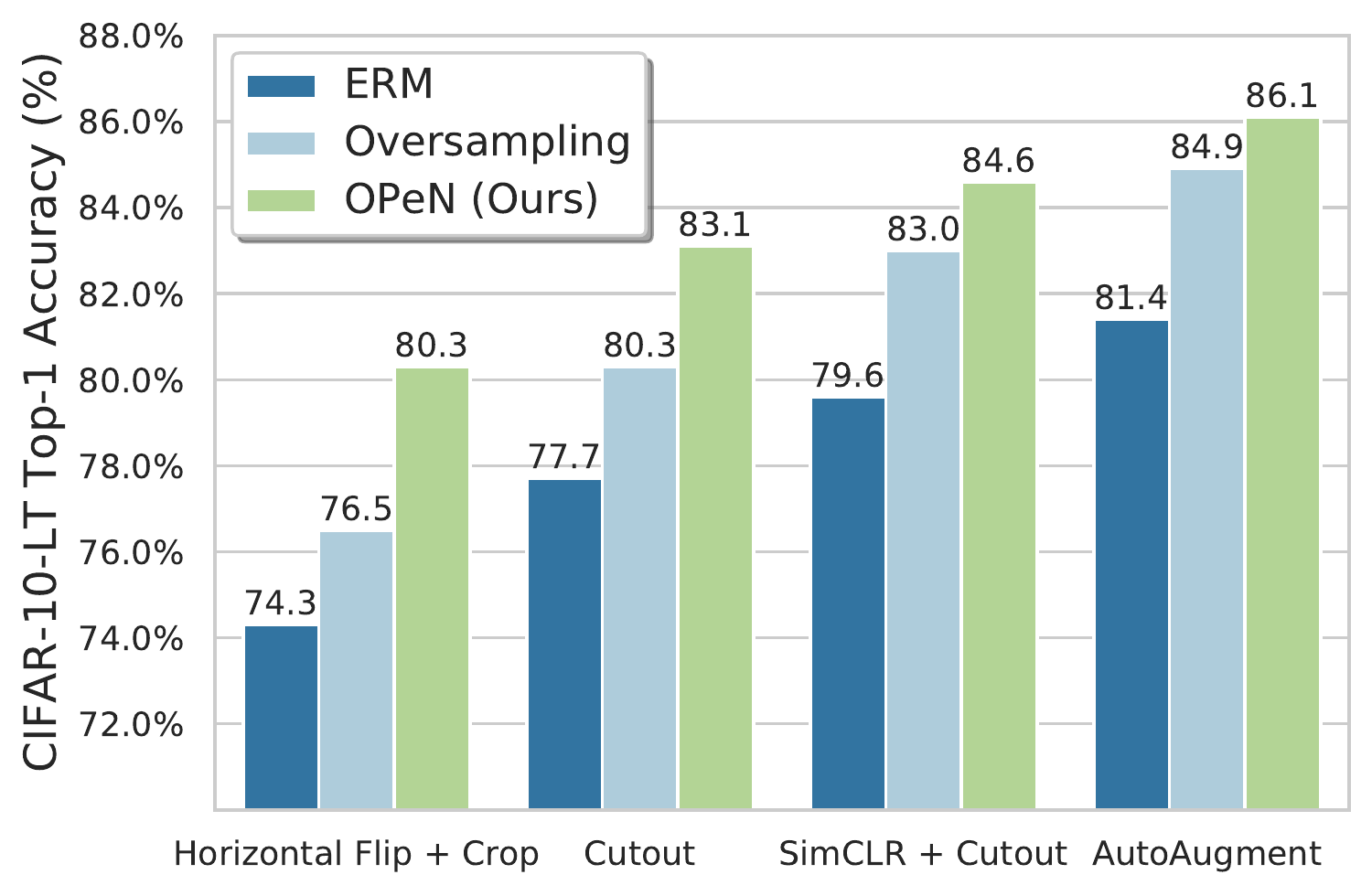}
    \vspace{-0.9cm}
	\caption{\textbf{Ablation study: The added value of pure noise w.r.t. various augmentation methods.} {\it Mean accuracy on  CIFAR-10-LT  with IR=100. 
	We compare OPeN to ERM baseline and to deferred oversampling~\cite{cao2019learning} (with same training parameters).
	}}
	\label{fig:augmentation_ablation}
	\vspace*{-0.55cm}
\end{figure}

\vspace{-0.1cm}
\section{Ablation Studies \& Observations}
\label{sec:ablation_study_observations}
In this section we explore the added-value of training on pure-noise images under various different settings, and the importance of using DAR-BN for batch normalization.

\vspace*{-0.1cm}
\noindent\textbf{Data augmentation.} 
In this ablation study, we explore the added value of pure noise when using OPeN with different types of data augmentation methods. We evaluate several augmentation techniques with increasing power on CIFAR-10-LT: (i) random horizontal flip, followed by random crop with padding of four pixels; (ii) Cutout~\cite{devries2017improved} (which zeros out a random fixed-size window in the image); (iii) SimCLR~\cite{chen2020simple} (which includes in addition to the horizontal flip and crop, also color distortion and Gaussian blur) followed by Cutout; (iv) AutoAugment~\cite{cubuk2019autoaugment} (which is optimized on the {entire balanced} CIFAR-10 and ImageNet datasets, and considered to be a highly-powerful augmentation). \cref{fig:augmentation_ablation} shows that OPeN provides a significant improvement over all four augmentation types, even when the optimal augmentation for that dataset (AutoAugment~\cite{cubuk2019autoaugment}) is used. This further supports our hypothesis that training on out-of-distribution pure noise images has a significant added value in suppressing overfitting, beyond augmentation of existing training images.

\noindent\textbf{The impact of DAR-BN.} 
Our distribution-aware normalization layer (DAR-BN) is an essential component for the success of our method, since it helps bridge the distribution gap between random pure noise images and natural images. 
\cite{xie2019intriguing, xie2020adversarial} already observed that natural images and adversarial images are drawn from two different domains.
They addressed this using an ``Auxiliary BN" layer, which separates  adversarial examples and clean images into two separate standard BN layers, with two separate learnable sets of affine parameters. In contrast, in DAR-BN we use only one set of trainable parameters, which are learned by activation maps of natural images only, and use them to scale and shift both the natural activation maps and the activation maps of the pure noise.
This normalization difference is important, since the test data in our case will  contain only natural images and no pure noise images.

\begin{table}
\vspace*{-0.1cm}
	\centering
	\resizebox{\linewidth}{!}{
	\begin{tabular}{@{}l|cc@{}}
		\toprule
		Norm Layer                              & CIFAR-10-LT                      & CIFAR-100-LT                     \\
		\midrule
		Standard BN{\footnotesize~\cite{ioffe2015batch}}       & $81.45${\scriptsize $\pm$0.70}                  & $49.18${\scriptsize $\pm$0.54}                   \\
		Auxiliary BN{\footnotesize ~\cite{xie2020adversarial}} & $83.38${\scriptsize $\pm$0.16}                   & $50.13${\scriptsize $\pm$0.06}                   \\
		\textbf{DAR-BN}                           & $\textbf{84.64}${\scriptsize $\pm$\textbf{0.16}} & $\textbf{51.50}${\scriptsize $\pm$\textbf{0.44}} \\
		\bottomrule
	\end{tabular}}
\vspace*{-0.4cm}
	\caption{\textbf{Ablation study: Comparing different Batch-Norm layers.} 
	{\it Mean accuracy on  CIFAR-10/100-LT with IR=100. Each type of BN is plugged into OPeN (with same training parameters).  DAR-BN outperforms the other normalization layers.}
	}
	\label{tab:ablation_normalization_layer}
	\vspace*{-0.6cm}
\end{table}

Table~\ref{tab:ablation_normalization_layer} compares the effect of plugging each of 3 different BN layers into OPeN:
(i)~Standard BN~\cite{ioffe2015batch}, (ii)~Auxiliary BN~\cite{xie2020adversarial}, and (iii)~DAR-BN (ours). Results show that DAR-BN outperforms other BN layers (surpassing  standard BN by 3.2\% and 2.3\% on CIFAR-10/100-LT, respectively).

\noindent\textbf{Pure Noise Images -- a General Useful Augmentation.}
Our method and experiments are primarily focused on imbalanced classification. However, we observed that adding pure noise images is often effective  as a \emph{general data enrichment method, which \underline{complements} existing augmentation methods}, even in standard balanced datasets.
To use it as such, we simply add 
a fixed number of pure noise images to each class (e.g., some pre-defined percentage of the class size), and train the network using DAR-BN as described in~\cref{alg:darbn}. We note that since our method does not modify existing training images, it can be easily applied in addition to any other augmentation technique.

While we did not perform extensive evaluations of this, we exemplify the potential power of training on pure-noise images (with DAR-BN) as an additional  useful data augmentation method, on the two \emph{full (balanced)} CIFAR datastes~\cite{krizhevsky2009learning}.
To examine the power of this ``complementary augmentation", 
we measure its added value \emph{on top of} 
successful and commonly used augmentation techniques: 
(i)~\textit{Baseline augmentation}: using random horizontal flip and random cropping (with crop size of 32 and padding of 4); (ii)~\textit{AutoAugment}~\cite{cubuk2019autoaugment} using the corresponding dataset policy; (iii)~\textit{Our method}: adding pure noise images (normalized with DAR-BN) in addition to AutoAugment.

We perform our experiments on the \emph{full (balanced)} CIFAR-10 and CIFAR-100, using a Wide-ResNet-28-10 architecture~\cite{zagoruyko2016wide}. All models were trained for 200 epochs using the Adam optimizer with $\beta_1=0.9$ and $\beta_2=0.999$, and with a standard cross-entropy loss. Noise images were sampled from a Gaussian distribution with mean and variance of the corresponding training set and with a noise-to-real ratio of 1:4 in each batch.
Our proposed method (AutoAugment complemented with pure-noise images) achieves the best classification accuracy on both datasets. Specifically:
\vspace{-0.1cm}
\begin{itemize}[topsep=0pt,itemsep=-1ex,partopsep=1ex,parsep=1ex,leftmargin=*]
\item{\textbf{Improvement over the baseline augmentation:}}\\
\hspace*{0.5cm} $\textbf{+2.38\%}$ on CIFAR-10, \ \ \  $\textbf{+6.34\%}$ on CIFAR-100.
\item{\textbf{Improvement over AutoAugment:}} \\
\hspace*{0.5cm} $\textbf{+0.9\%}$ on CIFAR-10, \ \ \ \  \  $\textbf{+1.5\%}$ on CIFAR-100. 
\end{itemize}
\vspace{-0.1cm}

\noindent 
These results suggest that 
properly utilizing pure noise images (with our proposed DAR-BN), may serve as an additional useful augmentation method in general, without any elaborated data creation schemes. It has the potential to further improve classification accuracy, even when used on top of highly sophisticated augmentation methods such as AutoAugment (which was optimized for these specific datasets).
Extensively verifying this observation on a large variety of datasets, architectures, and augmentation methods, is part of our future work.

\vspace{-0.1cm}
\section{Conclusion}
We present a new framework (OPeN) for imbalanced image classification: re-balance the training set by using pure noise images as additional training samples, along with a special distribution-aware normalization layer (DAR-BN). Our method achieves \emph{SOTA results on a large variety of imbalanced classification benchmarks}. In particular, it significantly improves generalization of tiny classes with very few training images. 
Our method is extremely simple to use, and can be incorporated in any training scheme. 
While we developed DAR-BN to bridge the distribution gap between real and pure-noise images, it may potentially serve as a new BN layer for bridging the gap between other pairs of different input domains in neural-nets.
Our work may open up new research directions for harnessing noise, as well as other types of out-of-distribution data, both for imbalanced classification, and for  data enrichment in general.

\textbf{Acknowledgement:} This project received funding from the D. Dan and Betty Kahn Foundation.

\bibliography{main}
\bibliographystyle{icml2022}

\newpage
\appendix
\onecolumn
\clearpage

\section{DAR-BN \& Distribution Shift}\label{appendix:darbn}
\begin{figure}[h!]
    \centering
    \begin{minipage}[c]{0.33\linewidth}
    \includegraphics[page=9,width=\textwidth]{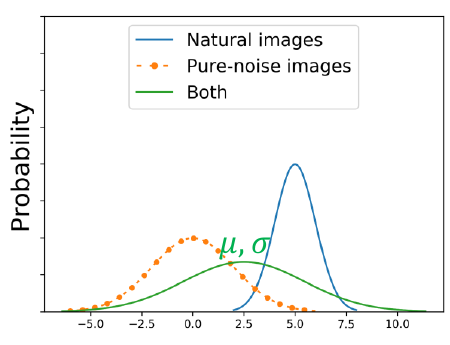}
    \end{minipage}
     \hfill
    \begin{minipage}[c]{0.33\linewidth}
    \includegraphics[page=10,width=\textwidth]{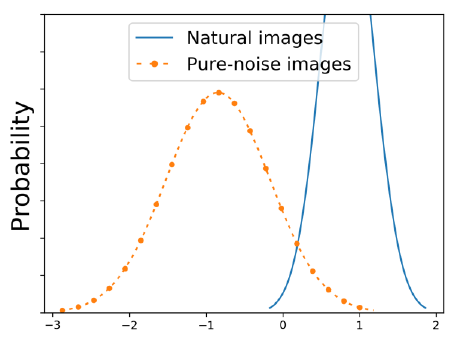}
    \end{minipage}
    \hfill
    \begin{minipage}[c]{0.33\linewidth}
    \includegraphics[page=8,width=\textwidth]{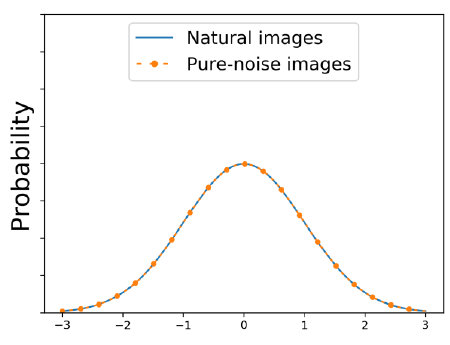}
    \end{minipage}
    \caption{\textbf{Illustration of normalization with BN versus DAR-BN}. {\it({\textbf{Left}})  Input of mixed natural (blue) and pure noise (orange) activation maps to be normalized. The input is assumed to be distributed as a bimodal normal distribution. ({\textbf{Center}}) Output of vanilla Batch Normalization layer which estimates only one set of mean and std (see the green line in the left figure) for both natural and noise activation maps. ({\textbf{Right}})  Output of DAR-BN layer that separately normalizes the natural and the noise activation maps. As required, the output is distributed as one Gaussian with zero mean and unit variance.}} \label{fig:distribution_shift_ilustraion}
\end{figure}

\begin{figure*}
\vspace{-0.3cm}
    \centering
    \begin{minipage}[c]{1\linewidth}
    \includegraphics[page=1,width=\textwidth]{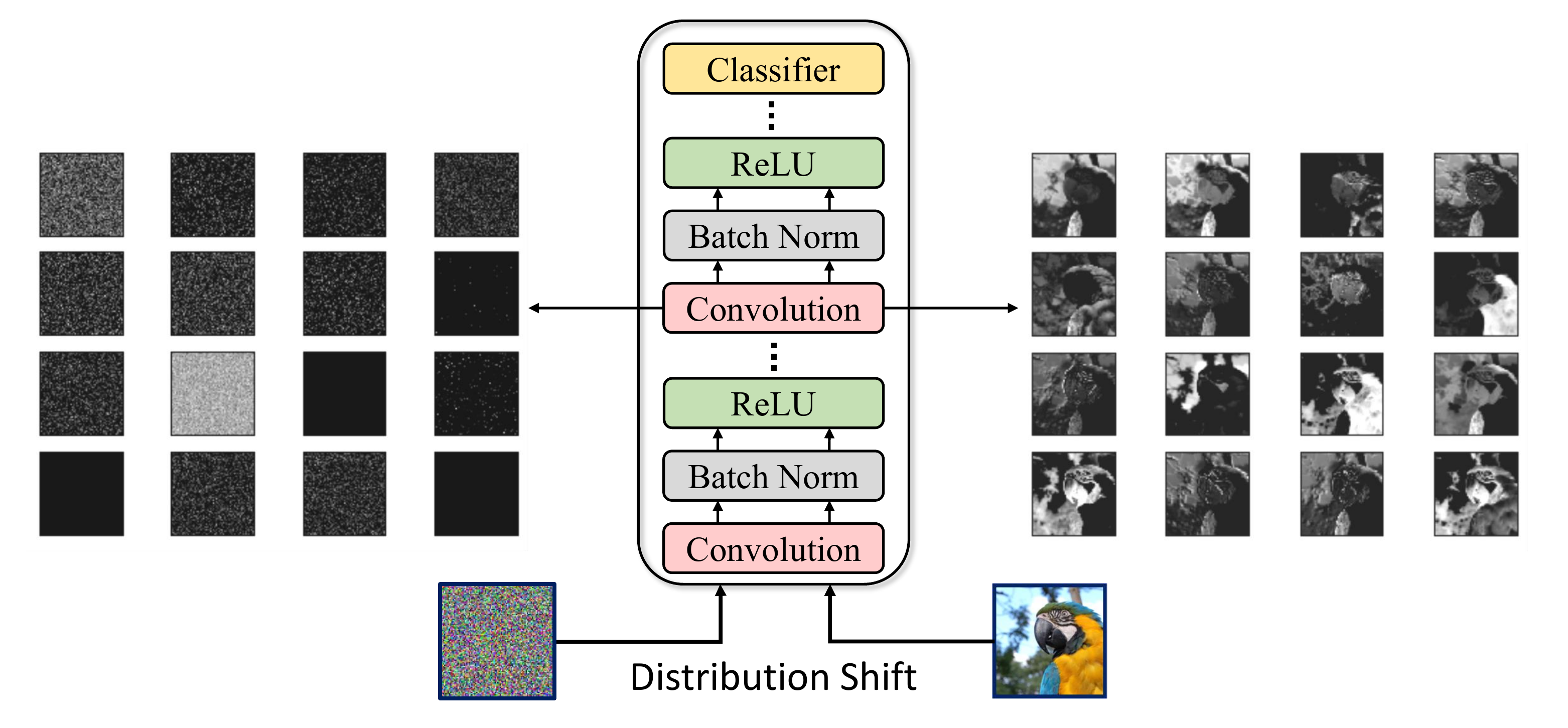}
    \end{minipage}
    \vspace{-0.25cm}
    \caption{\textbf{Distribution shift between natural images and noise images}. {\it
    Activation maps of a natural image and a pure random noise image. The activation maps are the output of the first convolution block of a pre-trained VGG16 model (we sample 16 out of 64 channels). We see that the features of the natural image and the pure-noise image were sampled from different distributions.}}\label{fig:distribution_shift_activation_map}
    \vspace{-0.15cm}
\end{figure*}

This section provides more details on Distribution-Aware Routing Batch Normalization (DAR-BN) layer discussed in~\cref{sec:dar_bn}. A detailed pseudo-code of DAR-BN layer is given in~\cref{alg:darbn}. In addition, we provide an illustration of the difference between the vanilla Batch Normalization and DAR-BN when the input comes from two different domains (see~\cref{fig:distribution_shift_ilustraion}).
While pure random noise is clearly out of the distribution of natural images, this distribution shift is kept also in deeper layers of the network (see~\cref{fig:distribution_shift_activation_map}).

\begin{algorithm}
	\caption{Distribution-Aware Routing BN (DAR-BN)}
	\begin{algorithmic}
		\STATE \textbf{Input:} (i) Batch of activation maps (per-channel) $\mathcal{X} = \set{x_j}_{j=1}^{batch_size}$ where $x_j$ is the channel activation map of example $j$;
		(ii) Function indicator $\mathbbm{1}_{noise}$ satisfies: $\mathbbm{1}_{noise}(x_j)=1 \Leftrightarrow x_j$ is an activation map of a pure noise image.
		\\\hrulefill
		\STATE Initialize $\gamma = 1$, $\beta = 0$, $\Bar{\mu}=0$, $\Bar{v}=0$
		\STATE splits $\leftarrow \set{'natural', 'noise'}$ 
		\FORALL {$split$ \textbf{in} $splits$}
		\STATE $is\_noise\_split\leftarrow$ $\mathbbm{1}_{\set{split == 'noise'}}$(split)
		\STATE $\triangleright$ {Split the batch}
		\STATE $\mathcal{X}_{split} \leftarrow \set{x\in \mathcal{X} | \mathbbm{1}_{noise}(x) == is\_noise\_split}$ 
		\STATE $\mu_{split} \leftarrow \mathbb{E}(\mathcal{X}_{split})$
		\STATE $v_{split} \leftarrow Var(\mathcal{X}_{split})$
		\STATE $\sigma_{split} \leftarrow \sqrt{v_{split}}$
		\STATE $\mathcal{X}_{split} \leftarrow \frac{\mathcal{X}_{split}-\mu_{split}}{\sigma_{split}}$
		\vspace{0.3em}
		\IF {$is\_noise\_split$}
		\STATE $\triangleright$ Do not update $\beta, \gamma$ as well as the batch statistics
		\STATE $\textbf{with}$ no gradient update: 
		\STATE \hspace{1.5em} $\mathcal{X}_{split}  \leftarrow \gamma\cdot \mathcal{X}_{split} -\beta$
		\ELSE
		\STATE $\mathcal{X}_{split}  \leftarrow \gamma\cdot \mathcal{X}_{split} -\beta$
        \STATE $\triangleright$  {Update statistics according to the natural split}
		\STATE $\Bar{\mu} \leftarrow \eta \cdot \mu_{split}   + (1-\eta) \cdot \Bar{\mu}$
		\STATE $\Bar{v} \leftarrow \eta \cdot v_{split}   + (1-\eta) \cdot \Bar{v}$
		\ENDIF
		\ENDFOR
	\end{algorithmic}
	\label{alg:darbn}
\end{algorithm}

\section{Using Additive Noise Instead of Pure-Noise Images}
\label{appendix:additive_noise}

\begin{figure}[h!]
    \centering
    \vspace*{-0.42cm}
    \includegraphics[page=1,width=0.93\textwidth]{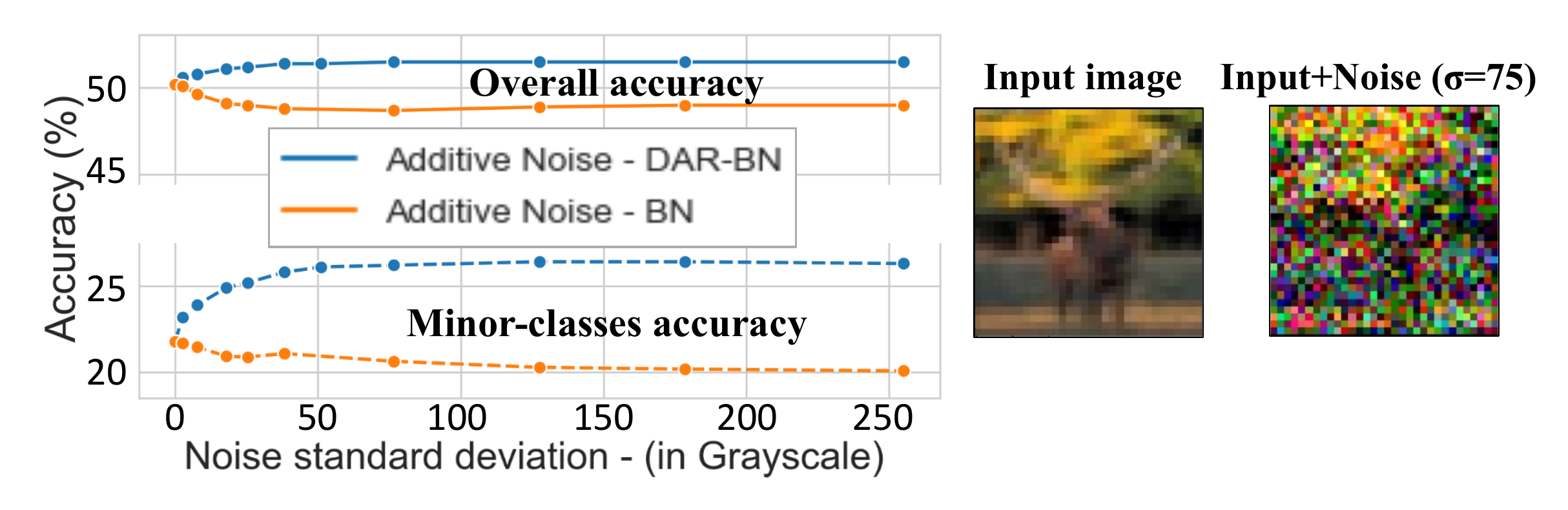}
    \vspace*{-0.42cm}
    \caption{\textbf{Additive Gaussian Noise Instead of Pure-Noise Images.} \textbf{Left:} The overall test accuracy (averaged over all classes - solid lines) and accuracy of the 20 smallest classes (dashed lines) of {CIFAR100-LT} (with IR=100, where minority classes have 5-12 samples). We use the same training scheme used by OPeN while using natural images with additive Gaussian noise instead of pure-noise images. Using DAR-BN results in significantly higher accuracy than vanilla BN. Best results are obtained where the additive noise practically becomes pure-noise (approximately where the standard deviation is higher than 75). \textbf{Right:} Example of a natural image and the same image with additive Gaussian noise with std=75.}
    \label{fig:pure_noise_verses_additive_noise}
\end{figure}

Using OPEN, we sample \emph{pure-noise} image from a normal distribution with fixed mean and standard deviation, determined by the mean and the standard deviation of the natural-image training set (\cref{sec:OPeN}). We experimented with replacing the \emph{pure-noise} images in OPeN with \emph{Additive Gaussian noise} of varying strengths (std) added to real images from the training set. Results on CIFAR-100-LT are plotted in \cref{fig:pure_noise_verses_additive_noise}, both with our DAR-BN (in blue) and vanilla BN (in orange). Each point on the graph relates to a model trained with a single noise strength (std) for all classes. Results show the increasing the strength of the added noise results in increased accuracy (in blue), until reaching a plateau at the point where the input image practically becomes pure noise ($\sigma \ge 75$), obtaining \emph{the same accuracy as our pure-noise images}. The most significant improvement is achieved for minor classes (blue dashed line), which is also maximized when approaching pure noise levels. This experiment demonstrates that adding training samples that are \emph{increasingly} `out-of-distribution' (i.e., increasingly larger additive noise) allows the model to better handle unseen test images of minor classes. This further supports our use of pure noise images. Note that when using vanilla BN instead of our DAR-BN, accuracy decreases when using higher std since the distribution shift becomes larger.

\section{\mbox{\bf {Generalization as a Function of the Class Size}}}\label{appendix:Generalization_class_size}
\vspace*{-0.15cm}
This section provides more details on the improvement provided by OPeN to the generalization of small (minority) classes (extending the evaluation in \cref{subsec:imbalanced_results}). To this goal, we perform a finer evaluation over the classes in CIFAR-10-LT and CIFAR-100-LT datasets. We divide the classes (according to their sample size) into five non-overlapping groups of equal size, i.e., each group consists of 20\% of the classes. For example, for CIFAR-100-LT, \textit{\text{Group \#1}} consists of the twenty smallest classes in the training set, while \textit{\text{Group \#5}} consists of the twenty largest classes. Similarly, for CIFAR-10-LT, each group consists of two classes.

\begin{figure}[h!]
  \centering
  \begin{minipage}[c]{0.48\linewidth}
    \includegraphics[page=1,width=\linewidth]{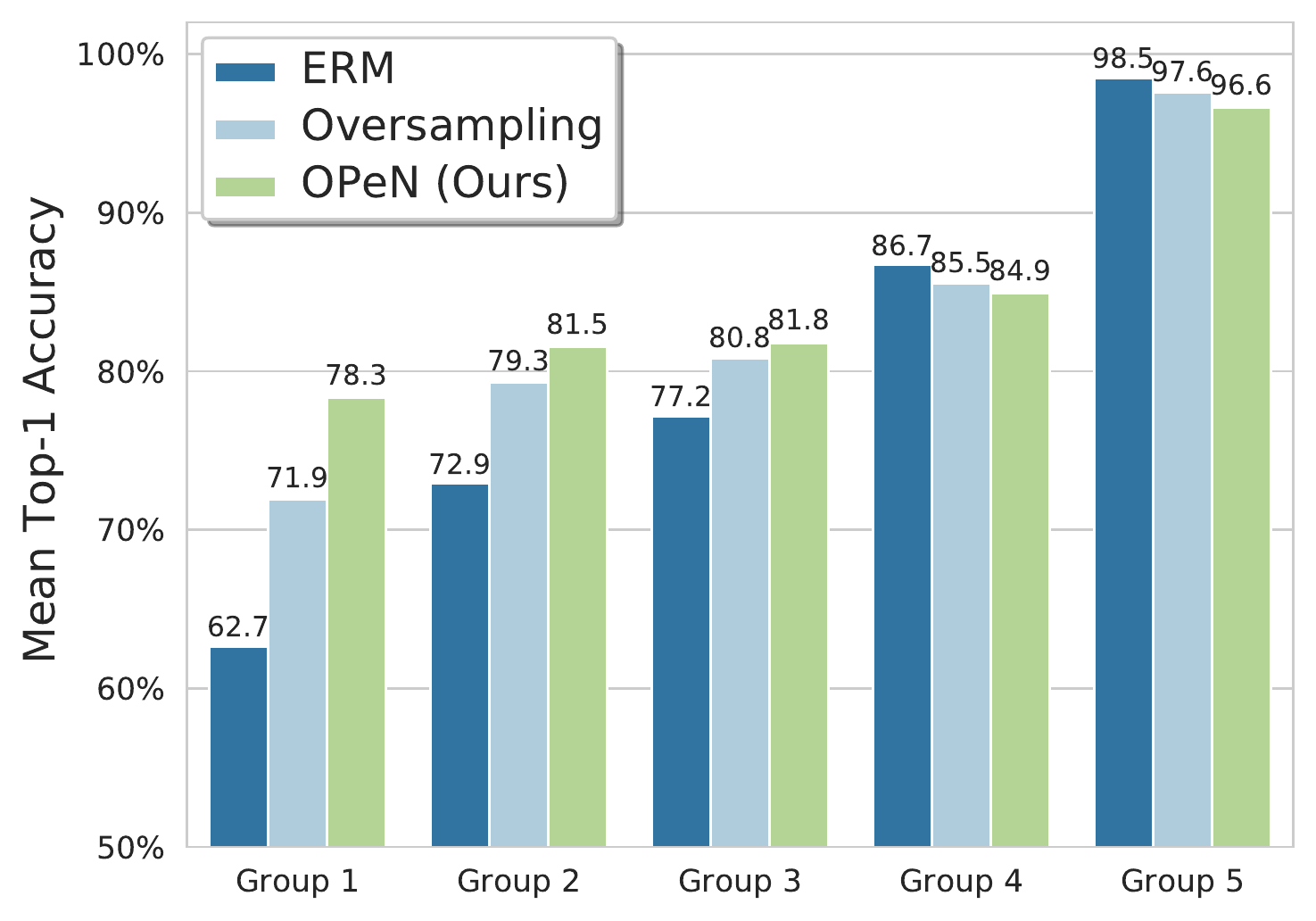}
    \centering
    \mbox{\hspace{0.8cm}(a) CIFAR-10-LT (IR=100)}
  \end{minipage}
  \begin{minipage}[c]{0.48\linewidth}
    \includegraphics[page=1,width=\linewidth]{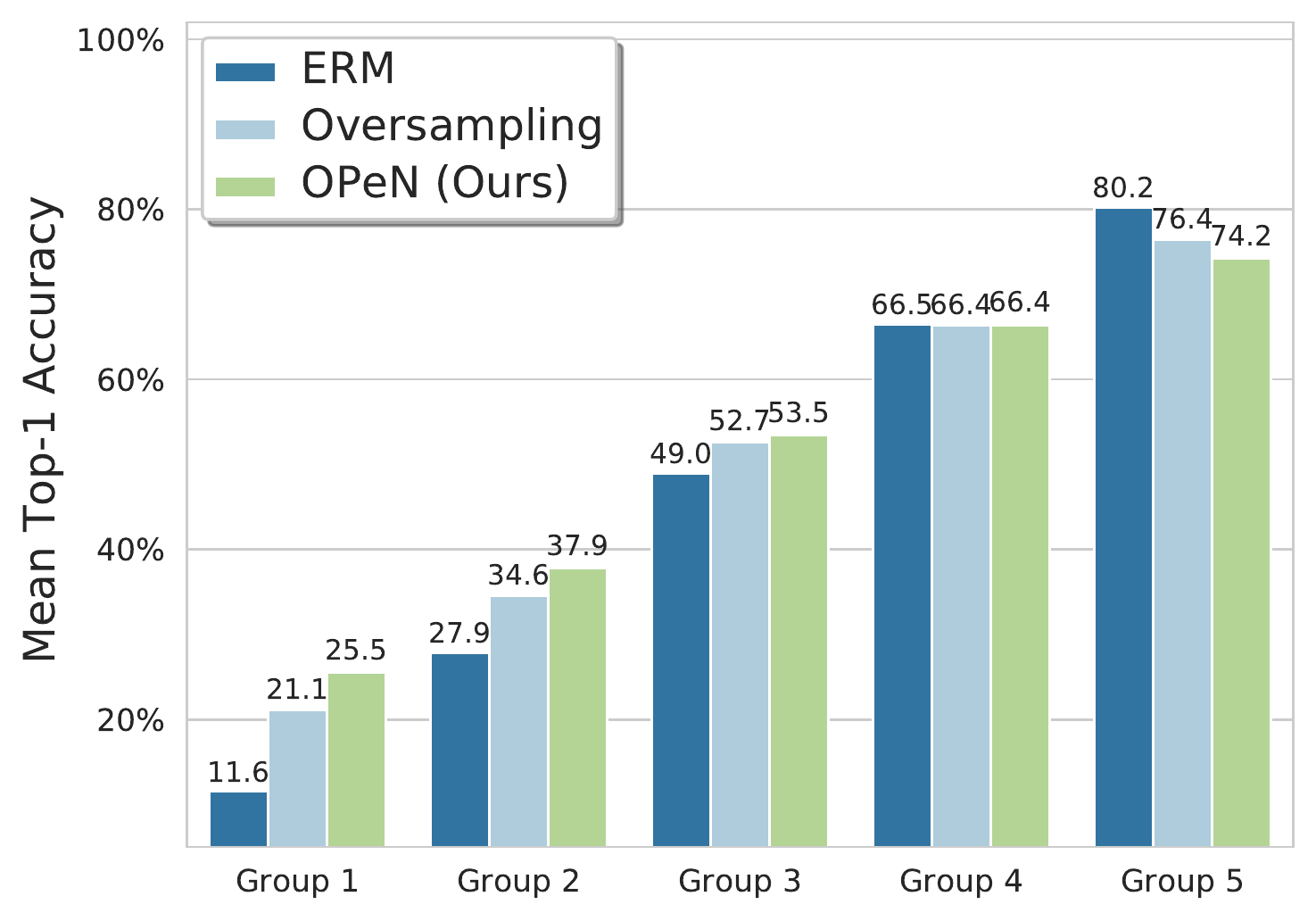}
    \centering
    \mbox{\hspace{0.8cm}(b) CIFAR-100-LT (IR=100)}
  \end{minipage}
  \vspace*{-0.1cm}
  \caption{\textbf{Mean accuracy for each group of classes.} \textit{We divide the classes into five groups according to their size and report the mean accuracy for each group of classes. Group 1 consists of the 20\% most minor classes, while Group 5 consists of the 20\% most major classes. We compare OPeN to both the baseline ERM and deferred oversampling~\cite{cao2019learning} (with the same training parameters). (a) Results on CIFAR-10-LT (IR=100), which has two classes in each group. (b) Results on CIFAR-100-LT (IR=100), which has twenty classes in each group. OPeN significantly improves the generalization of minority classes on both datasets.}}
  \label{fig:mean_accuracy_per_group}
\end{figure}

\Cref{fig:mean_accuracy_per_group} shows the classification results using OPeN compared to the ERM baseline and deferred oversampling~\cite{cao2019learning}, according to the classes division described above. OPeN provides a significant improvement over two methods on minority classes. We specifically note that \textbf{on CIFAR-10-LT, OPeN improves the accuracy over ERM for \textit{\text{Group \#1}} (the two smallest classes) by 15.6\%, for \textit{\text{Group \#2}} by 8.6\% and for \textit{\text{Group \#3}} by  4.6\% while degrades the accuracy for \textit{Group \#4 \& \#5} in less the 2\%}. This shows that using OPeN, besides improving the overall accuracy (discussed in \cref{subsec:imbalanced_results}), results in a more balanced classifier. These results support our claim that OPeN bypasses the limitation of using solely augmented images based on existing ones by employing out-of-distribution random images as additional training examples. Note that pure noise images add stochasticity to the training process. While this significantly improves generalization on severely and mildly overfitted minor classes (i.e., groups 1-3), noisy gradient may interfere with training on (non-overfitted) major classes (e.g., group 5).

\vspace{-0.30cm}

\vspace*{-0.35cm}
\section{Experimental Setup}\label{sec:Experimental_Setup}
\vspace*{-0.1cm}
\subsection{Architectures \& training}
\textbf{ImageNet-LT.} Following \cite{hong2021disentangling} We use ResNeXt-50-32x4d~\cite{xie2017aggregated} with a cosine classifier~\cite{gidaris2018dynamic}.We train it for 220 epochs using SGD optimizer with momentum 0.9 and weight decay \text{5e-4}. We use a step learning rate decay with an initial learning rate of \text{5e-2}, then decay by a factor of 0.1 at epochs 160 and 170. As in~\cite{cao2019learning,kim2020m2m}, we too defer our method (OPeN) to the last phase of the training. This allows the network to learn an initial representation of the data with natural images only. Only when the learning rate decays (which is when the model is exposed to overfitting), do we add the Oversampling + Pure-Noise images. For ImageNet OPeN is deferred to the last 40 epochs.

\textbf{Places-LT.} We follow the procedure of~\cite{hong2021disentangling,ren2020balanced,kang2019decoupling} which use ResNet152 pre-trained on ImageNet as a feature extractor. We use a randomly initialized cosine classifier on top of the backbone, and train the entire network end-to-end for additional 30 epochs using SGD optimizer with momentum 0.9 and weight decay \text{5e-4}. The initial learning-rate is set to \text{5e-2} for the classifier and \text{1e-3} for the backbone, then decay the learning rate by a factor of 0.1 at epochs 10 and 15 while OPeN is applied on epoch 15.

\textbf{CIFAR-10-LT, CIFAR-100-LT and CelebA-5.} All experiment use  WideResNet-28-10~\cite{zagoruyko2016wide}. For CIFAR-10/100-LT We follow the setup in ~\cite{kim2020m2m, cao2019learning}, that is, train for 200 epochs with Cross-Entropy loss using SGD optimizer with momentum 0.9 and weight decay of \text{2e-4}. We use a step learning rate decay with an initial learning rate of $0.1$, then decay by a factor of 0.01 at epochs 160 and 180. OPeN is deferred to the last 40 epochs. For CelebA-5\cite{kim2020m2m}, we follow \cite{kim2020m2m} and train for 90 epochs, decay the learning rate by a factor of 0.1 at epochs 30 and 60. 

The information of the five long-tailed datasets (ImageNet-LT, Places-LT, CIFAR-10-LT, \mbox{CIFAR-100-LT}, and CelebA-5) is summarized in \cref{tab:datasets_info}.

\textbf{Noise-ratio ($\delta$):} The noise-ratio $\delta$ was determined using the CIFAR validation set and set to $\frac{1}{3}$. We simply applied the same value to all other datasets.

\textbf{Randomization:} Since small datasets tend to present high variance, for CelebA-5, CIFAR-10-LT and CIFAR-100-LT, we repeat the experiments 4 times, reporting mean and standard error. For the rest of the datasets, for fair comparison, we use the same randomization seed across all experiments.

\begin{table*}
    \centering
    \resizebox{\linewidth}{!}{
    \begin{tabular}{@{}l|ccccc@{}}
        \toprule
        Dataset &
        \# of classes &
            Imbalance-ratio (IR) &
            Largest class size &
            Smallest class size &
            \# of samples  \\
        \midrule
        CIFAR-10-LT~\cite{cao2019learning}  & 10    & \{50\;,\;100\} & 5,000 & \{100\;,\;50\} & \{13,996\;,\;12,406\}     \\
        CIFAR-100-LT~\cite{cao2019learning} & 100   & \{50\;,\;100\} & 500   & \{10\;,\;5\}  & \{12,608\;,\;10,847\}     \\
        ImageNet-LT~\cite{liu2019large}           & 1,000 & 256 & 1,200 & 5  & 115,846  \\
        Places-LT~\cite{liu2019large}             & 365   & 996 & 4,980 & 5  & 62,500   \\
         CelebA-5 \cite{kim2020m2m}& 5 & 10.7  & 2423 &  227 & 6651  \\
  \bottomrule
    \end{tabular}}
    \vspace*{-0.2cm}
    \caption{\textbf{Long-tailed datasets.} {\it Summary of the long-tailed datasets we used for evaluation. 
    (see Sec.~\ref{sec:experiments} below for a detailed explanation).}}
    \label{tab:datasets_info}
\end{table*}

\subsection{Image Augmentations}
In each experiment, the following data augmentations were used:
\begin{itemize}[leftmargin=*]
    \item  For the datasets with small images (i.e., CIFAR-10-LT~\cite{cao2019learning}, CIFAR-100-LT~\cite{cao2019learning} and CelebA-5~\cite{kim2020m2m}), we use random horizontal flip followed by random crop with padding of four pixels, then apply Cutout~\cite{devries2017improved} (which zeros out a random $16\times 16$ window in the image) and SimCLR~\cite{chen2020simple} (which includes ColorJitter, random Grayscale and random GaussianBlur).
    \item  For the datasets with a higher resolution images (i.e., ImageNet-LT~\cite{liu2019large} and Places-LT~\cite{liu2019large}), we apply random resize crop (with default parameters) to $224\times224$ pixels followed by SimCLR and random rotation.
\end{itemize}

\noindent When AutoAugment~\cite{cubuk2019autoaugment} is employed (see \cref{sec:experiments}), it replaces all above augmentations. We use \text{CIFAR-10} policy for CIFAR-10-LT, CIFAR-100-LT and CelebA-5 datasets, and ImageNet policy for ImageNet-LT and Places-LT datasets.

\vspace{-0.3em}
\section{PyTorch Code}\label{sec:code}
In this section, we provide code snippets of the core components of our method.\\
Full code is available at \hyperlink{https://github.com/shiranzada/pure-noise}{https://github.com/shiranzada/pure-noise}.
\vspace{-0.7em}
\begin{itemize}
    \item \Cref{listing:DAR_BN} provides code implementation for the DAR-BN layer as described in the main paper. DAR-BN can be integrated into any neural network by replacing the standard BN layer. Specifically replace

\begin{lstlisting}[numbers=none]
x = self.batch_norm(x)
\end{lstlisting}
    with
\begin{lstlisting}[numbers=none]
x = dar_bn(self.batch_norm, x, noise_mask)
\end{lstlisting}
    where "noise\_mask" is a Boolean array that indicates which activation map is obtained from noise.
    
    \item \Cref{listing:OPeN} provides code implementation for OPeN together with pure noise images sampling function. The code is given in Python using the PyTorch library.
\end{itemize}

\begin{listing*}[h!]
\begin{minted}{python}

def dar_bn(bn_layer, x, noise_mask):
    """Applies DAR-BN normalization to a 4D input (a mini-batch of 2D inputs with 
        additional channel dimension)

    bn_layer : torch.nn.BatchNorm2d
        Batch norm layer operating on activation maps of natural images
    x : torch.FloatTensor of size: (N, C, H, W)
        2D activation maps obtained from both natural images and noise images
    noise_mask: torch.BoolTensor of size: (N)
        Boolean 1D tensor indicates which activation map is obtained from noise
    """
    # Batch norm for activation maps of natural images
    out_natural = bn_layer(x[torch.logical_not(noise_mask)])
    # Batch norm for activation maps of noise images
    # Do not compute gradients for this operation
    with torch.no_grad():
        adaptive_params = {"weight": bn_layer.weight, "bias": bn_layer.bias,
                           "eps": bn_layer.eps}
        out_noise = batch_norm_with_adaptive_parameters(x[noise_mask],
                                                        adaptive_params)
    # Concatenate activation maps in original order
    out = torch.empty_like(torch.cat([out_natural, out_noise], dim=0))
    out[torch.logical_not(noise_mask)] = out_natural
    out[noise_mask] = out_noise

    return out


def batch_norm_with_adaptive_parameters(x_noise, adaptive_parameters):
    """Applies batch normalization to x_noise according to adaptive_parameters

    x_noise : torch.FloatTensor of size: (N, C, H, W)
        2D activation maps obtained from noise images only
    adaptive_parameters:
        a dictionary containing:
            weight: scale parameter for the adaptive affine
            bias: bias parameter for the adaptive affine
            eps: a value added to the denominator for numerical stability.
    """
    # Calculate mean and variance for the noise activations batch per channel
    mean = x_noise.mean([0, 2, 3])
    var = x_noise.var([0, 2, 3], unbiased=False)
    # Normalize the noise activations batch per channel
    out = x_noise - mean[None, :, None, None]
    out = out / torch.sqrt(var[None, :, None, None] + adaptive_parameters["eps"])

    # Scale and shift using adaptive affine per channel
    out = out * adaptive_parameters["weight"][None, :, None, None]
          + adaptive_parameters["bias"][None, :, None, None]

    return out
\end{minted}
\caption{PyTorch code for DAR-BN layer.}
\label{listing:DAR_BN}
\end{listing*}

\begin{listing*}
\begin{minted}{python}

def oversampling_with_pure_noise_train_epoch(model, balanced_loader, criterion,
                                        optimizer, delta, num_samples_per_class,
                                        dataset_mean, dataset_std, image_size):
    """Trains model for one epoch according to the OPeN scheme

        model : torch.nn.Module;
            Model to train
        balanced_loader: torch.utils.data.DataLoader
            A class balanced loader - samples each class with equal probability
        delta: float
            Hyper-parameter for OPeN (see description in paper)
        num_samples_per_class: torch.IntTensor
            Number of samples in each class in the original imbalanced dataset
        dataset_mean: torch.FloatTensor of size: (3)
            Dataset mean per color channel
        dataset_std: torch.FloatTensor of size: (3)
            Dataset standard deviation per color channel
        image_size: int
            Image size - Assumes squared images of size (image_size , image_size)
    """
    for images, targets in balanced_loader:
        # Compute representation ratio
        max_class_size = torch.max(num_samples_per_class)
        representation_ratio = num_samples_per_class[targets] / max_class_size
        # Compute probabilities to replace natural images with pure noise images
        noise_probs = (1 - representation_ratio) * delta
        # Sample indexes to replace with noise according to Bernoulli distribution
        noise_indices = torch.nonzero(torch.bernoulli(noise_probs)).view(-1)
        # Replace natural images with sampled pure noise images
        noise_images = sample_noise_images(image_size=image_size, 
                    mean=dataset_mean, std=dataset_std, count=len(noise_indices))
        images[noise_indices] = noise_images
        # Create mask for noise images - later used by DAR-BN
        noise_mask = torch.zeros(images.size(0), dtype=torch.bool)
        noise_mask[noise_indices] = True
        # Train model
        outputs = model(images, noise_mask)
        loss = criterion(outputs, targets)
        optimizer.zero_grad()
        loss.backward()
        optimizer.step()
        

def sample_noise_images(image_size, mean, std, count):
    """Samples pure noise images from the normal distribution N(mean,std)"""
    r = torch.normal(mean[0], std[0], size=(count, 1, image_size, image_size))
    g = torch.normal(mean[1], std[1], size=(count, 1, image_size, image_size))
    b = torch.normal(mean[2], std[2], size=(count, 1, image_size, image_size))
    pure_noise_images = torch.cat((r, g, b), 1)

    return pure_noise_images
\end{minted}
\caption{PyTorch code for OPeN.}
\label{listing:OPeN}
\end{listing*}


\end{document}